\pdfoutput=1

\documentclass[11pt]{article}

\usepackage[]{ACL2023}

\usepackage{times}
\usepackage{latexsym}

\usepackage{url}            
\usepackage{booktabs}       
\usepackage{amsfonts}       
\usepackage{nicefrac}       
\usepackage{microtype}      
\usepackage{amsmath}
\usepackage{amssymb}
\usepackage{tablefootnote}
\usepackage{enumitem}
\usepackage{graphicx}       
\usepackage{caption}
\usepackage{subcaption}
\usepackage{pifont}
\usepackage{multirow}
\usepackage{makecell} 
\usepackage{ragged2e}
\usepackage{comment}
\usepackage{textgreek}
\usepackage{tipa}
\usepackage{lscape}

\usepackage{xspace}

\usepackage{todonotes}  
\definecolor{very-light-gray}{gray}{0.97}
\presetkeys{todonotes}{inline}{}
\presetkeys{todonotes}{size=footnotesize,backgroundcolor=very-light-gray}{}

\newcommand{\cmark}{\ding{51}}%
\newcommand{\xmark}{\ding{55}}%

\newcommand\red[1]{\textcolor{red}{#1}}
\newcommand\green[1]{\textcolor{green}{#1}}

\newcommand{\afriqa}{\textsc{AfriQA}\xspace}
\newcommand*{\yoruba}{Yor\`ub\'a\xspace}

\newcommand{\insertdatasetstatistics}{
\begin{table*}[t!]
    \small
    \setlength{\tabcolsep}{4pt}
    \begin{center}
    \begin{tabular}{lll|lll|ccc|r}
    \toprule[1pt]
    \textbf{Source}  & \multirow{2}{*}{\textbf{ISO}} & \textbf{Pivot}  & \textbf{African} & \multirow{2}{*}{\textbf{Script}}  & \textbf{\# Native} & \multirow{2}{*}{\textbf{Train}} & \multirow{2}{*}{\textbf{Dev}} & \multirow{2}{*}{\textbf{Test}} & \textbf{\% Unanswerable}  \\
    \textbf{Language} & & \textbf{Language} &  \textbf{Region} & & \textbf{Speakers} & & & & \textbf{Questions} \\
    \midrule[1pt]
    Bemba & bem & English & South, East \& Central & Latin  & 4M & 502  	& 503  	& 314 & 0.41 \\
    Fon & fon &  French  & West &  Latin & 2M & 427 & 428 & 386 & 0.22 \\
    Hausa & hau &  English & West & Latin  & 63M & 435  & 436  	& 300 & 0.36 \\
    Igbo & ibo &  English  & West & Latin  & 27M  & 417  & 418  	& 409 & 0.18 \\
    Kinyarwanda & kin &  English  & Central & Latin  & 15M & 407  & 409  & 347 & 0.26\\
    Swahili & swa &  English   & East \& Central & Latin  & 98M  & 415 & 417 & 302 & 0.34 \\
    Twi & twi &  English    & West & Latin  & 9M & 451  & 452  	& 490 & 0.12 \\
    Wolof & wol & French 	 &  West &  Latin & 5M & 503  & 504  & 334 & 0.38 \\
    \yoruba & yor &  English & West & Latin &  42M & 360 & 361 & 332 & 0.21 \\
    Zulu  & zul &   English	  & South &  Latin & 27M & 387  & 388  & 325 & 0.26 \\
    \midrule
    Total & --- & --- & --- &  --- & 292M & 4333  & 4346  & 3560 & 0.27 \\
    \bottomrule
    \end{tabular}
    \vspace{1em}
    \caption{\textbf{Dataset information:} This table contains key information about the \afriqa Dataset}
    \label{tab:datastats}
    \end{center}
    \end{table*} 
}

\newcommand{\insertretrievalexperimentstopten}{
\begin{table*}[!t]
     \begin{center}
     \small
    \begin{tabular}{l|ccc|cc|cc|cc|c}
    \toprule[1pt]
    & \multicolumn{3}{c|}{\textbf{Human Translation}} & \multicolumn{2}{c|}{\textbf{GMT}} & \multicolumn{2}{c|}{\textbf{NLLB}} & \multicolumn{2}{c|}{\textbf{M2M-100}} & \textbf{Crosslingual} \\
    \textbf{lang} & \textbf{BM25} & \textbf{mDPR} & \textbf{Hybrid} & \textbf{BM25} & \textbf{mDPR} & \textbf{BM25} & \textbf{mDPR} & \textbf{BM25} & \textbf{mDPR} & \textbf{mDPR} \\
    \midrule[1pt]
     & \multicolumn{10}{c}{Recall@10} \\
    \midrule[1pt]
    bem    	& 55.7  & 67.5  & \textbf{72.3}  & ---  & ---  & 52.2  & \textbf{59.8}  & ---  & --- & 14.7   \\
    fon  & 66.3 & 69.4 & \textbf{70.7} & --- & --- & 43.9 & \textbf{48.7} & 39.9 & 43.3 & 28.5	\\
    hau    	& 58.0  & 65.7  & \textbf{72.7}  & 53.3  & \textbf{60.3}  & 52.0  & 59.7  & 36.7 	& 44.3 & 13.7   \\
    igb    	& 70.4  & 74.3  & \textbf{82.9}  & 65.5  & \textbf{71.2}  & 64.8  & 68.0  & 62.1 & 67.5 & 25.4   \\
    kin    	& 59.1  & 66.3  & \textbf{75.5}  & 53.6  & \textbf{61.1}  & 53.0  & 58.8  & --- 	&  --- & 15.6  \\
    swa   & 46.0 & 61.9 & \textbf{67.6} & 45.0  & \textbf{60.9} & 43.1 & 58.3 & 39.1 & 54.6 & 20.9	\\
    twi    	& 61.8  & 66.7  & \textbf{75.3}  & 56.1  & \textbf{58.0}  & 50.4  & 54.1  &  45.7	& 49.4   & 21.4   \\
    wol    	& 61.4  & 67.7  & \textbf{68.6}  & ---  & --- &  35.0 &  	\textbf{36.5} & 34.4 	 & 35.0  & 13.8  \\
    yor   & 55.1 & 66.6	& \textbf{71.7} & 52.1 & \textbf{59.0} & 50.9 & 57.5 & 36.8 & 35.5 & 21.4  \\
    zul    	& 59.7  & 70.2  & \textbf{76.3}  & 57.2  & \textbf{66.2}  & 51.5  & 64.6  & 45.5  & 60.0  &  14.2 \\
    \midrule
    avg & 59.4 & 67.6 & \textbf{73.4} & 54.7 & \textbf{62.4} & 49.7 & 56.6 & 42.5 & 48.7 & 19.0 \\
    \midrule[1pt]
    & \multicolumn{10}{c}{Recall@100} \\
    \midrule
    bem	& 76.8 	& 81.9 	& \textbf{84.7}  & ---  & --- 	& 70.4 	& \textbf{74.2} 	&  ---	 & --- & 37.3     \\
    fon	& 78.8 	& 79.3 & \textbf{80.1} & --- & --- & \textbf{60.3} & 59.3 & 59.6 & 59.3 & 46.9 \\
    hau & 77.7 	& 83.3 	& \textbf{84.7}  & 77.7  & \textbf{79.3} 	& 75.0  & 77.7 	& 58.3 	 & 64.3 & 34.3     \\
    igb & 87.0 	& 89.7 	& \textbf{94.6}  & 85.6 & \textbf{87.5}  & 84.8 	& 83.9  &  82.4	 & 83.4  & 50.1     \\
    kin & 78.1  & 81.3 	& \textbf{87.0} & 75.2	& \textbf{78.1}	& 74.1	& 77.0	& --- & ---  & 30.3     \\
    swa &  70.9 & 80.5 & \textbf{82.1} & 68.1 & \textbf{79.8} & 68.2 & 77.2 & 64.2 & 76.2 & 40.1 \\
    twi & 78.4  & 82.9  & \textbf{85.7} & 71.6 & \textbf{83.7}  & 70.0  & 72.5  & 61.8 &  63.1 & 38.4     \\
    wol	& 82.6 	& 82.6 & \textbf{84.7}  & --- & --- & \textbf{56.0} &  55.1	& 57.2 & 53.6 & 31.1   \\
    yor &  78.6 & 83.4 & \textbf{87.1} & 73.2 & \textbf{79.2} & 71.1 & 78.3 & 59.6 & 55.4 & 46.7 \\
    zul & 86.2  & 86.2 	& \textbf{91.1}  & \textbf{83.1}  & 72.0 	& 77.0	& 80.6 	&  71.1	 &  74.8	&  28.9   \\
    \midrule
    avg & 79.5 & 83.1 & \textbf{86.2} & 76.4 & \textbf{79.9} & 70.8 & 73.6  & 64.3 & 66.3 & 38.4  \\
    \bottomrule[1pt]
    \end{tabular}
    \vspace{1em}
    \caption{\textbf{Retrieval Recall@10/100}: This table displays the retrieval recall results for various translation types on the test set of \afriqa. The table shows the percentage of retrieved passages that contain the answer for the top-10 and top-100 retrieved passages. The last column represents crosslingual retrieval, where we skip the translation step and use the original queries. We boldface the best-performing model for each language within the human translation oracle scenario and within the real-world automatic translation scenario. 
    }
    \label{tab:retrievalexperimentstopten}
    \end{center}
    \end{table*} 
}

\newcommand{\insertretrievalexperimentstoptwenty}{
\begin{table*}[!t]
     \begin{center}
     \small
    \begin{tabular}{l|ccc|cc|cc|cc|c}
    \toprule
    & \multicolumn{3}{c|}{Human Translation} & \multicolumn{2}{c|}{GMT} & \multicolumn{2}{c|}{NLLB} & \multicolumn{2}{c|}{M2M-100} & Multilingual \\
    & BM25 & mDPR & Hybrid & BM25 & mDPR & BM25 & mDPR & BM25 & mDPR & mDPR \\
    \midrule
    \midrule
    lang & \multicolumn{10}{c}{\textbf{Recall@20}} \\
    \midrule
    bem  & 64.3  & 72.6  & \textbf{76.8}  & ---  & ---  & 60.2  & 65.3  & ---  & ---  & 22.0     \\
    fon  & 71.5 & 72.2	& \textbf{74.6} & --- & --- & 49.6 & 52.3 & 46.5 & 46.9 & 30.3 \\
    hau  & 64.3  & 73.3  & \textbf{78.0}  & 60.0  & 70.0  & 59.3  & 68.7  & 43.3 & 51.7  & 20.0     \\
    igb  & 75.3  & 78.7  & \textbf{87.8}  & 72.4  & 76.0  & 70.2  & 73.4  & 67.2 & 74.3  & 34.0     \\
    kin  & 67.4  & 72.6  & \textbf{80.1}  & 63.1  & 68.6  & 62.0  & 65.7  & --- &  --- & 19.3     \\
    swa  & 54.6 & 67.6 & \textbf{72.5} & 52.7 & 66.9 & 50.3 & 64.6 & 47.0 & 61.3 & 26.8 \\
    twi  & 69.0  & 71.4  & 7\textbf{8.4}  & 61.0  & 63.7  & 55.9  & 58.6  & 49.8 & 53.9 & 26.3     \\
    wol  & 68.6  & \textbf{73.1}  & 72.2  & ---  & ---  &42.8  & 43.7 	&  41.0	 &  40.4 & 18.0     \\
    yor  & 62.7 & 72.6 & \textbf{77.7} & 58.4 & 66.9 & 58.1 & 65.7 & 41.9 & 41.9 & 31.3 \\
    zul  & 68.6  & 76.6  & \textbf{83.7}  & 66.5  & 71.7  & 62.2  & 69.2  & 53.2  & 64.9 & 18.2    \\
    \bottomrule
    \end{tabular}
    \vspace{1em}
    \caption{\textbf{Retrieval recall@20}: This table presents the retrieval recall@20 results for different translation types on the test set of \afriqa. This shows the percentage of the top 20 retrieved passages that contain the answer. Multilingual retrieval skips the translation step}
    \label{tab:retrievalexperimentstoptwenty}
    \end{center}
    \end{table*} 
}

\newcommand{\insertpivotlanguagespan}{
\begin{table*}[!t]
     \begin{center}
     \small
    \begin{tabular}{r|l|cccccccccc|c}
    \toprule
     &  & \multicolumn{11}{c}{Pivot Language Span F1} \\
    \midrule
     \textbf{Query Translation} & \multicolumn{1}{c|}{\textbf{Retrieval}} & \textbf{bem} & \textbf{fon} & \textbf{hau} & \textbf{ibo} & \textbf{kin} & \textbf{swa} & \textbf{twi} & \textbf{wol} & \textbf{yor} & \textbf{zul} & avg   \\
    \midrule
    HT & BM25 & 29.2 & \textbf{11.4}  & 31.4 & 43.0 & 33.8 & 24.3 & 38.4 & 15.4  & 28.9 & 32.8 & 28.9  \\
    HT & mDPR & 32.5 & 11.0 & \textbf{35.8} & 44.8 & 35.4 & 28.2 & 40.7 & 14.7 & 31.7 & \textbf{36.5} & 31.1  \\
    HT & Hybrid & \textbf{34.7} & 11.3 & 35.5 & \textbf{46.1} & \textbf{39.2} & 27.5 & \textbf{41.8} & \textbf{16.2} & \textbf{32.4} & 34.6 & \textbf{32.0}  \\
    GMT & BM25 & --- & --- & 21.0 & 38.6 & 28.3 & 24.7 & 27.7 & --- & 21.7 & 31.6 & 27.7 \\
    GMT & mDPR & --- & --- & 31.5 & 39.3 & 35.3 & \textbf{29.1} & 31.1 & --- & 22.9 & 36.0 & 32.2  \\
    NLLB & BM25 & 23.8 & 3.6 & 24.6 & 37.6 & 29.3 & 25.2 & 25.7 & 4.4 & 17.3 & 26.8 & 19.8 \\
    NLLB & mDPR & 24.1 & 5.1 & 27.2 & 39.6 & 33.3 & 25.9 &28.2 & 5.2 & 21.4 & 30.4 & 24.0 \\
    \midrule
     &  & \multicolumn{11}{c}{Pivot Language Span EM} \\
    \midrule
    HT & BM25 & 21.4 &\textbf{8.0}   & 24.0 & 31.1 & 17.3 & 17.5 & 25.3 & 10.2 & 21.4 & 23.1 & 19.9  \\
    HT & mDPR & 23.2 & 7.0 & 26.7 & 32.5 & 19.3 & \textbf{20.9} & 27.6 & 10.8  & 22.9 & 24.0 & 21.5  \\
    HT & Hybrid & \textbf{25.2} & 7.3 & 26.3 & \textbf{33.3} & \textbf{22.2} & 19.5 & \textbf{28.2} & \textbf{11.1} & \textbf{23.2} & 23.1 & \textbf{21.9}  \\
    GMT & BM25 & --- & --- & \textbf{27.8} & 30.3 & 16.1 & 18.2 & 17.8 & --- & 16.6 & 21.5 & 21.2 \\
    GMT & mDPR & --- & --- & 22.7 & 30.1 & 20.7 & 20.5 & 20.4 & --- & 16.6 & \textbf{24.9} & 22.3 \\
    NLLB & BM25 & 14.6 & 0.8 & 19.0 & 28.9 & 15.6 & 18.5 & 15.9 & 3.3 & 12.7 & 19.1 & 13.8 \\
    NLLB & mDPR & 14.3 & 2.1 & 20.7 & 29.1 & 18.7 & 18.9 & 18.2 & 2.7 & 14.5 & 20.6 & 16.0 \\
    \bottomrule
    \end{tabular}
    \vspace{1em}
    \caption{F1  and EM scores on pivot language answer generation using an extractive multilingual reader model with different query translation and retrieval methods.}
    \label{tab:pivot_lang_span_retrieve}
    \end{center}
    \end{table*} 
}

\newcommand{\insertxorfull}{
\begin{table*}[!t]
     \begin{center}
     \small
     \setlength{\tabcolsep}{5pt}
    \begin{tabular}{cc|l|cccccccccc|ccc}
    \toprule
    \multicolumn{2}{c|}{Translation} &  & & \\
    Query & Answer & \multicolumn{1}{c|}{Retrieval} &  \multicolumn{10}{c|}{XOR-Full F1} & \multicolumn{3}{c}{Average} \\
    \midrule
     &  &  & \textbf{bem} & \textbf{fon} & \textbf{hau} & \textbf{ibo} & \textbf{kin} & \textbf{swa} & \textbf{twi} & \textbf{wol} & \textbf{yor} & \textbf{zul} & \textbf{F1} & \textbf{EM} & \textbf{BLEU}   \\
    \midrule
    GMT & GMT & BM25 & --- & --- & 20.4 & 30.4 & 24.2 & 18.1 & 14.9 & --- & 16.1 & 19.7 & 20.5 & 12.1 & 18.3  \\
    GMT & GMT & mDPR & --- & ---  & \textbf{21.7} & \textbf{33.0} & \textbf{26.5} & \textbf{21.9} & 16.5 & 14.2 & \textbf{20.4} & \textbf{21.1} & \textbf{23.0} & \textbf{14.2} & \textbf{20.7} \\
    NLLB & NLLB & BM25 & \textbf{13.6} & 2.6 & 17.5 & 26.5 & 19.9 & 19.2 & 18.4 & 3.2 & 12.7 & 12.5 & 14.6 & 7.5 & 12.9 \\
    NLLB & NLLB & mDPR  & 13.3 & 4.3 & 19.3 & 29.9 & 22.4 & 20.3 & \textbf{19.5} & \textbf{3.5} & 17.6 & 13.1 & 16.3 & 8.3 & 14.3\\
    \midrule
    \end{tabular}
    \vspace{1em}
    \caption{XOR-Full F1 results combining different translation and retriever systems.}
    \label{tab:fulltrip}
    \end{center}
    \end{table*} 
}

\newcommand{\insertxorfullem}{
\begin{table*}[!t]
     \begin{center}
     \small
     \setlength{\tabcolsep}{5pt}
    \begin{tabular}{cc|l|cccccccccc|cc}
    \toprule
    \multicolumn{2}{c|}{Translation} &  & & \\
    Query & Answer & \multicolumn{1}{c|}{Retrieval} &  \multicolumn{10}{c|}{XOR-Full BLEU} & \multicolumn{1}{c}{Average} \\
    \midrule
     &  &  & \textbf{bem} & \textbf{fon} & \textbf{hau} & \textbf{ibo} & \textbf{kin} & \textbf{swa} & \textbf{twi} & \textbf{wol} & \textbf{yor} & \textbf{zul} & \textbf{BLEU}   \\
    \midrule
    GMT & GMT & BM25 & --- & --- & 19.4 & 28.2 & 21.1 & 16.0 & 11.7 & --- & 13.8 & 18.1 & 18.3 \\
    GMT & GMT & mDPR & --- & --- & 20.1 & 30.3 & 23.3 & 19.9 & 13.2 & --- & 18.6 & 19.6 &\textbf{20.7} \\
    NLLB & NLLB & BM25 & 11.4 & 1.7 & 15.9 & 24.8 & 16.8 & 16.9 & 16.6 & 2.9 & 10.9 & 10.7 & 12.9 \\
    NLLB & NLLB & mDPR & 10.9 & 3.3 & 17.0 & 27.2 & 18.8 & 18.3 & 17.5 & 3.1 & 15.3 & 11.4 & 14.3 \\
    \midrule
    & & & \multicolumn{10}{c|}{XOR-Full EM} & \textbf{EM} \\
     \midrule
    GMT & GMT & BM25 & --- & --- & 16.3 & 21.0 & 12.3 & 10.9 & 4.0 & --- & 8.0 & 12.0 & 12.1 \\
    GMT & GMT & mDPR & --- & --- & 15.7 & 22.7 & 15.0 & 14.6 & 4.9 & --- & 12.7 & 14.2 & \textbf{14.2} \\
    NLLB & NLLB & BM25 & 6.7 &0.5  & 11.7 & 15.4 & 7.8 & 10.0 & 10.6 & 2.4 & 5.1 & 4.3 & 7.5 \\
    NLLB & NLLB & mDPR  & 5.4 & 0.2 & 10.7 & 17.6 & 8.6 & 15.3 & 10.8 & 2.4  & 7.2 & 4.9 & 8.3\\
    \midrule
    \end{tabular}
    \vspace{1em}
    \caption{XOR-Full results }
    \label{tab:fulltripembleu}
    \end{center}
    \end{table*} 
}

\newcommand{\insertMTblue}{
\begin{table}[!t]
     \begin{center}
     \small
     \setlength{\tabcolsep}{3pt}
    \begin{tabular}{llccc}
    \toprule
    \textbf{Source} & \textbf{Target} & \multirow{2}{*}{\textbf{GMT}}  &  \multirow{2}{*}{\textbf{NLLB}}  & \multirow{2}{*}{ \textbf{M2M-100}}  \\
    \textbf{lang} & \textbf{lang} \\
    \midrule
    bem & eng & ---  & \textbf{24.4} & --- \\
    fon & fre & --- & \textbf{16.6} & 8.7 \\
    hau & eng & \textbf{55.2} & 44.6 & 26.3  \\
    ibo & eng & \textbf{48.3} & 46.3 & 34.1  \\
    kin & eng & \textbf{44.9} & 43.1 & ---  \\
    swa & eng & \textbf{54.0} & 53.2 & 34.7 \\
    twi & eng & \textbf{33.0} & 30.1 & 15.7 \\
    wol & fre & --- & \textbf{16.6} & 12.7 \\
    yor & eng & \textbf{32.7} & 30.6 & 10.6 \\
    zul & eng & \textbf{50.2} & 45.4  & 33.3  \\
    \midrule
    avg & --- & \textbf{45.5} & 35.1 & 22.0\\
    \bottomrule
    \end{tabular}
    \vspace{1em}
    \caption{\textbf{Translation BLEU Scores:} BLEU score of some translation systems on the test set for the answer translation task. Note that Google Translate is not yet available in all languages, due to their very low-resource nature.}
    \label{tab:mt_bleu}
    \vspace{-1em}
    \end{center}
    \end{table} 
}

\newcommand{\insertSampledData}{

\begin{table*}[!t]
     \begin{center}
     \small
     \renewcommand{\arraystretch}{1.4}
     \resizebox{\textwidth}{!}{
    \begin{tabular}{lllcc}
    \toprule
    \textbf{lang} & \textbf{Question $Q_L$}  & \textbf{Relevant Passage $P_{pl}$}  &  \textbf{Answer $A_{pl}$} \\
    & (\emph{Translation $Q_{pl}$}) & & (\emph{Translation $A_L$}) \\    
    \midrule
    \midrule
    \multirow{3}{*}{hau} &  \multirow{3}{*}{\makecell[l]{Jahohi nawa ne a kasar Malaysia? \\ banga? (\emph{How many states are there} \\ \emph{in Malaysia?})}}  & \multirow{3}{*}{\makecell[l]{The states and federal territories of Malaysia are the principal \\ administrative divisions of Malaysia. Malaysia is a  federation \\ of {\color[HTML]{14AA34} \textbf{13}} states (Negeri) and 3 federal territories.}} & \multirow{3}{*}{13 (\emph{13})}  \\
    &  &   & \\
    &  &  & \\
    \midrule
    \multirow{3}{*}{bem} & \multirow{3}{*}{\makecell[l]{Bushe Mwanawasa stadium ingisha \\ abantu banga? (\emph{What is the capacity of} \\  \emph{Mwanawasa Stadium?})}} & \multirow{3}{*}{\makecell[l]{The Levy Mwanawasa Stadium is a multi-purpose stadium in \\ Ndola, Zambia. It is used mostly for football matches. The \\stadium has a capacity of {\color[HTML]{14AA34} \textbf{49,800 people}.} }} & \multirow{3}{*}{ \makecell{49,800 people \\ (\emph{Abantu 49800})} }  \\
    &  &  & \\
    & &  & \\
    \midrule
    \multirow{3}{*}{wol} & \multirow{3}{*}{\makecell[l]{Man po moo niroo ag powum Softbal? \\ (\emph{Quel sport ressemble beaucoup } \\  \emph{au softball?})}} & \multirow{3}{*}{\makecell[l]{Ce sport est un descendant direct du {\color[HTML]{14AA34} \textbf{baseball}} (afin de \\ différencier les deux) mais diffère de ce dernier par différents  \\ aspects dont les cinq principaux sont les suivants.}} & \multirow{3}{*}{ \makecell{baseball \\ (\emph{Bas-bal})} }  \\
    &  &  & \\
    & &  & \\
    \midrule
    \multirow{4}{*}{zul} & \multirow{3}{*}{\makecell[l]{Kwenzeka ngamuphi unyaka \\ uMlilo Omkhulu waseLondon? \\ (\emph{In what year did the Great Fire } \\ \emph{of London occur? })}} & \multirow{4}{*}{\makecell[l]{Great Fire of London The Great Fire of London was a major \\ conflagration that swept through the central parts of London \\ from Sunday 2 September to Thursday, 6 September {\color[HTML]{14AA34} \textbf{1666}} \\ The fire gutted the medieval City of London inside the wall.}} & \multirow{3}{*}{ \makecell{1666 \\ (\emph{1666})} }  \\
    &  &  & \\
    & &  & \\
    & &  & \\
    \bottomrule
    \end{tabular}
    }
    \vspace{1em}
    \caption{Table showing selected questions, relevant passages, and answers in different languages from the dataset. It also includes the human-translated versions of both questions and answers. For the primary XOR QA task, systems are expected to find the relevant passage among all Wikipedia passages, not simply the gold passage shown above.}
    \label{tab:dataset_examples}
    \end{center}
    \end{table*} 
}

\newcommand{\insertpivotlangspan}{
\begin{table}[!t]
    \setlength{\tabcolsep}{3.5pt}
     \begin{center}
     \small
    \begin{tabular}{l|cc|cc|cc|cc}
    \toprule
    & \multicolumn{2}{c|}{HT} & \multicolumn{2}{c|}{GMT} & \multicolumn{2}{c|}{NLLB} & \multicolumn{2}{c}{Crosslingual} \\
    &F1 & EM & F1 & EM & F1 & EM & F1 & EM\\
    \midrule
    bem & \textbf{38.2} & \textbf{29.5} & --- & --- & 30.0 & 21.9 & 0.4 & 0.4   \\
    fon & \textbf{53.8} & \textbf{40.4} & --- & --- & 37.5 & 26.7 & 13.4 & 6.0 \\
    hau & \textbf{60.9} & \textbf{52.7} & 54.4 & 47.7 & 50.9 & 43.7 & 27.7 & 23.7   \\
    ibo & \textbf{68.2} & \textbf{60.6} & 62.1 & 55.0 & 62.8 & 56.2 & 29.2 & 24.7 \\
    kin & \textbf{56.8} & \textbf{38.9} & 50.8 & 36.0 & 51.3 & 36.6 & 22.7 & 17.9 \\
    swa & \textbf{45.2} & \textbf{37.9} & 44.6 & 37.9 & 45.2 & 38.1 & 31.6 & 24.6   \\
    twi & \textbf{51.2} & \textbf{41.8}&39.2&31.1&34.3&30.0&3.4 & 2.5 \\
    wol & \textbf{45.2} & \textbf{33.9} & --- & --- & 33.2 & 26.0 & 1.8 & 0.9 \\
    yor & \textbf{45.1} & \textbf{38.6} & 36.0 & 31.7 & 32.3 & 28.0 & 6.0 & 3.8 \\
    zul & \textbf{59.1} & \textbf{49.2} & 56.0 & 48.6 & 53.6 & 45.8 & 17.0 & 13.5  \\
    \midrule
    avg & \textbf{52.4} & \textbf{42.4} & 42.9 & 36.0 & 43.1 & 35.3 & 15.3 & 11.8 \\
    \bottomrule
    \end{tabular}
    \vspace{1em}
    \caption{\textbf{ Extractive Gold Passages Answer Prediction:} Comparison of F1 and Exact Match Accuracy scores for extractive answer span prediction on the test set using AfroXLMR-base~\cite{alabi2022adapting} as the backbone.}
    \label{tab:pivotlangspanextractive}
    \end{center}
    \end{table} 
}

\newcommand{\insertpivotlangspanmt}{
\begin{table}[!t]
    \setlength{\tabcolsep}{3.5pt}
     \begin{center}
     \small
    \begin{tabular}{lcc|cc|cc|cc}
    \toprule
    & \multicolumn{2}{c|}{HT} & \multicolumn{2}{c|}{GMT} & \multicolumn{2}{c|}{NLLB} & \multicolumn{2}{c}{Crosslingual} \\
    &F1 & EM & F1 & EM & F1 & EM & F1 & EM \\
    \midrule
    bem & \textbf{48.8} & \textbf{41.7} & --- & --- & 38.5 & 32.0 & 2.9 & 1.1  \\
    fon & \textbf{41.4} & \textbf{28.5} & --- & --- & 23.4 & 15.3 & 5.1 & 2.3 \\
    hau & \textbf{58.5} & \textbf{49.0} & 53.5 & 45.7 & 50.9 & 42.7 & 25.8 & 22.3 \\
    ibo & \textbf{66.6} & \textbf{59.2} & 59.8 & 53.3 & 60.2 & 53.3 & 41.7 & 34.7 \\
    kin & \textbf{60.8} & \textbf{43.8} & 57.3 & 40.9 & 58.8 & 42.9 & 25.5 & 20.2 \\
    swa & \textbf{52.3} & \textbf{42.6} & 48.9 & 40.8 & 49.2 & 41.2 & 29.4 & 23.5  \\
    twi & \textbf{55.4} & \textbf{45.3} & 42.0 & 33.7 & 40.1 & 33.1 & 5.3 & 3.5 \\
    wol & \textbf{44.6} & \textbf{36.1} & --- & --- & 21.8 & 16.9 & 3.9 & 2.8  \\
    yor & \textbf{54.9} & \textbf{49.8} & 48.9 & 45.1 & 47.9 & 43.0 & 11.9 & 7.8 \\
    zul & \textbf{60.2} & \textbf{50.8} & 57.4 & 48.9 & 55.6 & 46.5 & 24.7 & 20.9 \\
    \midrule
    avg & \textbf{54.5} & \textbf{44.7} & 46.0 & 38.6 & 44.6 & 36.7 & 17.6 & 13.9 \\
    \bottomrule
    \end{tabular}
    \vspace{1em}
    \caption{\textbf{Generative Gold Passages Answer Prediction:} Comparison of F1 and Exact Match Accuracy scores for generative answer span prediction on the test set using mT5-base~\cite{xue2020mt5} as the backbone.}
    \label{tab:pivotlangspanmt5}
    \end{center}
    \end{table} 
}

\newcommand{\insertdatasetcomparison}{
    \begin{table*}[!t]
    \footnotesize
    \setlength{\tabcolsep}{4pt}
    \begin{center}
    \begin{tabular}{lccccl}
    \toprule
    Dataset & \textbf{QA?}& \textbf{CLIR?} & \textbf{Open Retrieval?} & \textbf{\# Languages} & \textbf{\# African Languages}  \\ 
    \toprule
    XQA \cite{liu-etal-2019-xqa} & \green{\cmark} & \green{\cmark} & \green{\cmark} & 9 & Nil\\
    XOR QA \cite{asai-etal-2021-xor} & \green{\cmark} & \green{\cmark} & \green{\cmark} & 7 & Nil \\
    XQuAD \cite{artetxe-etal-2020-cross} & \green{\cmark} & \red{\xmark} & \red{\xmark} & 11 & Nil \\
    MLQA \cite{lewis-etal-2020-mlqa} & \green{\cmark} & \red{\xmark} & \red{\xmark} & 7 & Nil  \\
    MKQA \cite{longpre-etal-2021-mkqa} & \green{\cmark} & \red{\xmark} & \green{\cmark} & 26 & Nil  \\
    TyDi QA \cite{clark-etal-2020-tydi} & \green{\cmark} & \red{\xmark} & \green{\cmark} & 11 & 1   \\
     AmQA \cite{abedissa2023amqa} & \green{\cmark} & \red{\xmark} & \red{\xmark} & 1 & 1   \\
      KenSwQuAD \cite{10.1145/3578553} & \green{\cmark} & \red{\xmark} & \red{\xmark} & 1 & 1   \\
    \midrule
     \afriqa (Ours) & \green{\cmark} & \green{\cmark} & \green{\cmark} &  10 & 10 (see \autoref{tab:datastats}) \\
    \toprule
    \end{tabular}
    \caption{\textbf{Comparison of the Dataset with Other Question Answering Datasets.} This table provides a comparison of the current dataset used in the study with other related datasets. The first, second, and third columns, ``QA'', ``CLIR'', and ``Open Retrieval'', indicate whether the dataset is question answering, cross-lingual or open retrieval, respectively. The fourth column, "\# Languages", shows the total number of languages in the dataset. The final column lists the African languages present in the dataset. 
    }
    \label{tab:corpuscomparison}
    \end{center}
    \end{table*}
}

\newcommand{\inserttrainingparams}{
    \begin{table}[!t]
    \small
    \setlength{\tabcolsep}{4pt}
    \begin{center}
    \begin{tabular}{lc}
    \toprule
    Parameters & Value \\
    \toprule
    backbone & multilingual-bert\\
    \# train epochs & 25 \\
    \# warmup steps & 500 \\
    \# GPUs & 4 \\
    \# gradient accumulation & 2 \\
    learning rate &  5.0e-05 \\
    $\epsilon$ & 1.0e-08 \\ 
    batch size & 16 \\
    weight decay & 0.01 \\
    max gradient norm & 1.0 \\
    seed & 42 \\
    max sequence length & 256 \\
    \toprule
    \end{tabular}
    \caption{\textbf{DPR Reader Training Configurations}}
    \label{tab:modelconfig}
    \end{center}
    \end{table}
}

\newcommand{\insertlanginfo}{
\begin{landscape}
\begin{table}[p!]
     \begin{center}
     \small
     \renewcommand{\arraystretch}{1.5}
     \setlength{\tabcolsep}{3pt}
     \setlength\extrarowheight{-3.0pt}
    \begin{tabular}{ll|lllll}
    \toprule
    \textbf{Language} & \textbf{Family} & \multirow{2}{*}{\makecell[l]{\textbf{Morphological} \\ \textbf{Inflection}}} & \textbf{Tenses} & \textbf{Negation} & \textbf{Plurality} & \textbf{WH-questions} \\
    & & & & \\
    \midrule[1pt] 
    \multirow{4}{*}{\textbf{Bemba}} & \multirow{4}{*}{Niger--Congo} & \multirow{4}{*}{Very rich} & \multirow{4}{*}{\makecell[l]{Affix to head word \\present ``ali'', past ``aali''}} & \multirow{4}{*}{\makecell[l]{Affix to head word: \\ “ta”, “shi”, and “kaana” }} & \multirow{4}{*}{\makecell[l]{Affix to the steam \\ of the word depending\\ on noun class }} & \multirow{4}{*}{\makecell[l]{What: ``cinshi'', Who: ``naani'' \\ When: ``liisa'', Why: ``mulandunshi''\\ Which: ``ciisa'', Where: ``kwi/kwiisa'', \\ How: “shaani”   }} \\
    & \\
    & \\
    & \\
    \midrule
    \multirow{3}{*}{\textbf{Fon}} & \multirow{3}{*}{Niger--Congo} & \multirow{3}{*}{Little or none} & \multirow{3}{*}{\makecell[l]{New word added:  \\past ``xóxó''}} & \multirow{3}{*}{\makecell[l]{New word added: ``ǎ''}} & \multirow{3}{*}{\makecell[l]{New word added: “l\'\textepsilon”}} & \multirow{3}{*}{\makecell[l]{What:``Et\'\textepsilon'',  Who: ``M\textepsilon'' \\ When: ``Hwet\'\textepsilon nu'', Why: ``Aniw\'{u}'' \\ Which: ``\textrtaild e t\textepsilon \ '', Where: ``Fit\textepsilon'' }} \\
    & \\
    & \\
    \midrule
    \multirow{3}{*}{\textbf{Hausa}} & \multirow{3}{*}{Afro--Asiatic} & \multirow{3}{*}{Rich} & \multirow{3}{*}{\makecell[l]{Indicative form Words used to \\indicate tenses: past: ``tsohon'' (was) \\ present: ``yanzu'' (is)}}  & \multirow{3}{*}{\makecell[l]{Indicative form. Words used to \\ indicate negation:  ba/ba a'' (not) \\ and ``banda'' (except)}} & \multirow{3}{*}{\makecell[l]{Suffix with vowel deletion. E.g.: \\ ``hula'' (cap), ``huluna'' (caps) \\ ``mace'' (girl), ``mataye'' (girls)}} & \multirow{4}{*}{\makecell[l]{What: ``me/ya'’, Who: ``wa'’ \\ When: ``yaushe'', Why: ``dan me/akan me'' \\ Which: ``wanne'', Where: ``ina/ a ina'' \\ How: ``yaya/qaqa''}} \\
    & \\
    & \\
    & \\
    \midrule
    \multirow{3}{*}{\textbf{Igbo}} & \multirow{3}{*}{Niger--Congo} & \multirow{3}{*}{Rich} & \multirow{3}{*}{None} & \multirow{3}{*}{Suffix “ghi”} & \multirow{3}{*}{\makecell[l]{No suffix. Count is often specified \\after the word}} & \multirow{4}{*}{\makecell[l]{What: kedu/gini, Who: onye/kedu onye \\ When: kedu mgbe, Why: gini mere/gini \\ Which: kedu nke, Where: ebee \\ How: kedu ka or kedu etu}} \\
    & \\
    & \\
    & \\
    \midrule
    \multirow{3}{*}{\textbf{Kinyarwanda}} & \multirow{3}{*}{Niger--Congo} & \multirow{3}{*}{Very rich} & \multirow{3}{*}{\makecell[l]{Changes to morphemes\\in a word}} & \multirow{3}{*}{\makecell[l]{Changes to morphemes\\in a word}} & \multirow{3}{*}{\makecell[l]{Changes to morphemes\\in a word}} & \multirow{3}{*}{\makecell[l]{What: ``iki'', Who: ``nde/inde'' \\ When: “ryari”, Which: ``ikihe/uwuhe'' \\ Where: ``hehe'', How: ``gute'' }} \\
    & \\
    & \\
    \midrule
    \multirow{3}{*}{\textbf{Swahili}} & \multirow{3}{*}{Niger--Congo} & \multirow{3}{*}{Very rich} & \multirow{3}{*}{\makecell[l]{Present: ``ni'' (is), \\Past: ``alikuwa'' (was/former)\\ Future: ``atakuwa'' (will be)}} &\multirow{3}{*}{---} & \multirow{3}{*}{\makecell[l]{Indicated by changes to the \\prefix according to noun class}} & \multirow{3}{*}{\makecell[l]{What: ``nii'', Who: ``nani'', When: ``lini'' \\ Why: ``kwanini'', Which: ``upi'', \\ Where: ``upi'', How: ``vipi” }}\\
    & \\
    & \\
    \midrule
    \multirow{3}{*}{\textbf{Twi}} & \multirow{3}{*}{Niger--Congo} & \multirow{3}{*}{Rich} & \multirow{3}{*}{None} & \multirow{3}{*}{\makecell[l]{``n'' is added to the root word}} & \multirow{3}{*}{\makecell[l]{Indicated by replacing the first \\ two letters of a root word with \\ ``mm'' or ``nn''.}} & \multirow{3}{*}{\makecell[l]{What: ``\textepsilon de\textepsilon n'', Who: ``hwan'', \\ When: daben, Why: adεn, Which: de\textepsilon hen \\Where: \textepsilon henfa, How: s\textepsilon n }}\\
    & \\
    & \\
    \midrule
    \multirow{3}{*}{\textbf{Wolof}} & \multirow{3}{*}{Niger--Congo} & \multirow{3}{*}{Rich} & \multirow{3}{*}{\makecell[l]{Dependent word: past tense, \\``oon'' is attached to the end \\ of the verb}} & \multirow{3}{*}{\makecell[l]{Keyword ``ul'' is added at the end \\ of the verb e.g nekk -- > nekkul}} & \multirow{3}{*}{\makecell[l]{Dependent word: plurality, ``yi''\\ or ``ay'' is attached before or after \\ the word}} & \multirow{3}{*}{\makecell[l]{What: ian, Who: kan, When: kañ \\ Why: lu tax, Which: ban, \\ Where: fan, How: naka}}\\
    & \\
    & \\
    \midrule
    \multirow{3}{*}{\textbf{Yoruba}} & \multirow{3}{*}{Niger--Congo} & \multirow{3}{*}{Little or none} & \multirow{3}{*}{\makecell[l]{To indicate present tense, keyword ``n''.\\ Past tense is indicated with ``ti'' with or
 \\ without a time period}} & \multirow{3}{*}{\makecell[l]{Keywords such as \textit{k\`{o}}. \\ \textit{máa}, \textit{nile}}} & \multirow{3}{*}{\makecell[l]{Count is specified with a word}} & \multirow{4}{*}{\makecell[l]{What: “Kini”, Who: “Tani” \\ When: “iga / nigba”, Why: “kilode” \\ Which: “ewo”, Where: “Nibo” \\ How is: “bawo”, How many: “elo / meloo”}}\\
    & \\
    & \\
    & \\
    \midrule
    \multirow{3}{*}{\textbf{Zulu}} & \multirow{3}{*}{Niger--Congo} & \multirow{3}{*}{Very rich} & \multirow{3}{*}{\makecell[l]{Present: affix after subject concord \\  (e.g.  ``ya'' or ``sa'') \\ Past: suffix (e.g.  ``e'' or ``ile'') }} & \multirow{3}{*}{\makecell[l]{Typically indicated by \\ the prefix  ``nga-}} & \multirow{3}{*}{\makecell[l]{Indicated by morphemes \\ ``aba'', ``izi'',``imi'',``o''}} & \multirow{3}{*}{\makecell[l]{What: ``yini”, Who: ``ubani'', When: “nini” \\ Why: “kungani”, Which: “yiliphi”, \\ Where: “kuphi”, How: “kanjani”}}\\
    & \\
    & \\
    \bottomrule[1pt]
    \end{tabular}
    \vspace{1em}
    \caption{\textbf{Language Linguistic Features:} This table provides a breakdown of the typologies, grammatical structures, and phonology of the 10 languages in \afriqa}
    \label{tab:languagelinguistic}
    \end{center}
\end{table} 
\end{landscape}
}

\usepackage[T1]{fontenc}

\usepackage[utf8]{inputenc}

\usepackage{inconsolata}

    \makeatletter
\def\@fnsymbol#1{\ensuremath{\ifcase#1\or \dagger\or \ddagger\or
   \mathsection\or \mathparagraph\or \|\or **\or \dagger\dagger
   \or \ddagger\ddagger \else\@ctrerr\fi}}
    \makeatother

\title{AfriQA: Cross-lingual Open-Retrieval\\Question Answering for African Languages}

\author{
\normalsize Odunayo Ogundepo$^{1,*,}$\thanks{\hspace{0.1cm} Equal contribution. We list detailed contributions in \textsection \ref{sec:contributions}.} , Tajuddeen R. Gwadabe$^{*,}$\footnotemark[1] , Clara E. Rivera$^{2}$, Jonathan H. Clark$^{2}$, Sebastian Ruder$^{2}$, \\
\textbf{\normalsize David Ifeoluwa Adelani$^{3,*}$, Bonaventure F. P. Dossou$^{4,5,6,*}$, Abdou Aziz DIOP$^{7,*}$, Claytone Sikasote$^{10,*}$, }  \\
\textbf{\normalsize  Gilles Hacheme$^{9,*}$, Happy Buzaaba$^{15,*}$, Ignatius Ezeani$^{14,*}$, Rooweither Mabuya$^{16}$, Salomey Osei$^{*}$, } \\
\textbf{\normalsize  Chris Emezue$^{13,*}$, Albert Njoroge Kahira$^{17,*}$, Shamsuddeen Hassan Muhammad$^{18,*}$, Akintunde Oladipo$^{1,*}$,} \\
\textbf{\normalsize  Abraham Toluwase Owodunni$^{*}$, Atnafu Lambebo Tonja$^{12,*}$,  Iyanuoluwa Shode$^{11,*}$, Akari Asai$^{8}$,} \\
\textbf{\normalsize  Tunde Oluwaseyi Ajayi$^{19,*}$,Clemencia Siro$^{20,*}$, Steven Arthur$^{21,*}$, Mofetoluwa Adeyemi$^{1,*}$, } \\
\textbf{\normalsize Orevaoghene Ahia$^{8,*}$, Aremu Anuoluwapo$^{*}$, Oyinkansola Awosan$^{*}$, Chiamaka Chukwuneke$^{*}$} \\
\textbf{\normalsize Bernard Opoku$^{22,*}$, Awokoya Ayodele$^{23,*}$, Verrah Otiende$^{24,*}$, Christine Mwase$^{25,*}$, Boyd Sinkala$^{10,*}$} \\
\textbf{\normalsize Andre  Niyongabo Rubungo$^{26,*}$, Daniel A. Ajisafe$^{27,*}$, Emeka Felix  Onwuegbuzia$^{23,*}$, Habib  Mbow$^{28,*}$  } \\
\textbf{\normalsize Emile Niyomutabazi$^{29,*}$, Eunice Mukonde$^{10,*}$, Falalu Ibrahim Lawan$^{30,*}$, Ibrahim Said Ahmad$^{31,*}$, } \\
\textbf{\normalsize Jesujoba O. Alabi$^{32,*}$, Martin Namukombo$^{33,*}$, Mbonu Chinedu$^{35,*}$, Mofya Phiri$^{10,*}$, Neo  Putini$^{25,*}$,} \\
\textbf{\normalsize Ndumiso  Mngoma$^{31,*}$, Priscilla  A.  Amuok$^{*}$, Ruqayya Nasir Iro$^{32,*}$, Sonia Adhiambo$^{34,*}$} \\
\footnotesize
$^*$Masakhane NLP, $^1$University of Waterloo, Canada, $^2$Google Research,$^3$University College London, \\
\footnotesize
$^4$Mila Quebec AI Institute,$^{5}$Research Center of Intelligent Machines, McGill University, $^{6}$Lelapa AI, $^7$GalsenAI,\\
\footnotesize
$^9$Ai4Innovr, $^8$University of Washington, $^{10}$ University of Zambia, $^{11}$Montclair State University, \\
\footnotesize
 $^{12}$Instituto Politécnico Nacional, Mexico, $^{13}$Technical University of Munich, $^{14}$Lancaster University, \\
 \footnotesize
 $^{15}$RIKEN Center for AIP, $^{16}$South African Centre for Digital Language Resources, $^{17}$Jülich Supercomputing Centre, \\
 \footnotesize
 $^{18}$University of Porto,$^{19}$Insight Centre for Data Analytics, $^{20}$University of Amsterdam, $^{21}$Accra Institute of Technology, \\
 \footnotesize
 $^{22}$Kwame Nkrumah University of Science and Technology,$^{23}$University of Ibadan, $^{24}$Tom Mboya University \\
 \footnotesize
 $^{25}$Fudan University,$^{26}$University of Electronic Science and Technology of China,  $^{27}$University of British Columbia, \\
 \footnotesize
  $^{28}$African Master in Machine Intelligence,$^{29}$College de Rebero, $^{30}$Kaduna State University, $^{31}$Bayero University Kano \\
\footnotesize
 $^{32}$Saarland University,  Germany,$^{33}$University of Edinburgh,$^{34}$ Kenyatta University,$^{35}$Nnamdi Azikiwe University  \\  
}

\begin{document}

\maketitle

\begin{abstract}
African languages have far less in-language content available digitally, making it challenging for question answering systems to satisfy the information needs of users. Cross-lingual open-retrieval question answering (XOR QA) systems---those that retrieve answer content from other languages while serving people in their native language---offer a means of filling this gap. To this end, we create \afriqa, the first cross-lingual QA dataset with a focus on African languages. \afriqa includes 12,000+ XOR QA examples across 10 African languages. While previous datasets have focused primarily on languages where cross-lingual QA \textit{augments} coverage from the target language, \afriqa focuses on languages where cross-lingual answer content is the \textit{only} high-coverage source of answer content. Because of this, we argue that African languages are one of the most important and realistic use cases for XOR QA. Our experiments demonstrate the poor performance of automatic translation and multilingual retrieval methods. Overall, \afriqa proves challenging for state-of-the-art QA models. We hope that the dataset enables the development of more equitable QA technology.\footnote{The data is available at: \\ 
\href{https://github.com/masakhane-io/afriqa}{https://github.com/masakhane-io/afriqa}.}

\end{abstract}

\section{Introduction}

\insertdatasetcomparison

Question Answering (QA) systems provide access to information \cite{kwiatkowski2019natural} and increase accessibility in a range of domains, from healthcare and health emergencies such as COVID-19 \cite{moller-etal-2020-covid,morales2021covid} to legal queries \cite{martinez2021survey} and financial questions \cite{chen-etal-2021-finqa}. Many of these applications are particularly important in regions where information and services may be less accessible and where language technology may thus help to reduce the burden on the existing system.
At the same time, many people prefer to access information in their local languages---or simply do not speak a language supported by current language technologies \cite{amano2016languages}.
To benefit the more than three billion speakers of under-represented languages around the world, it is thus crucial to enable the development of QA technology in local languages.

Standard QA datasets mainly focus on English \cite{joshi-etal-2017-triviaqa,mihaylov-etal-2018-suit,kwiatkowski2019natural,sap-etal-2020-commonsense}. While some reading comprehension datasets are available in other high-resource languages \cite{ruder2021multi}, only a few QA datasets \cite{clark-etal-2020-tydi,asai-etal-2021-xor,longpre-etal-2021-mkqa} cover a typologically diverse set of languages---and very few datasets include African languages (see~\autoref{tab:corpuscomparison}).

In this work, we lay the foundation for research on QA systems for one of the most linguistically diverse regions by creating \afriqa, the first QA dataset for 10 African languages. \afriqa focuses on open-retrieval QA where information-seeking questions\footnote{These questions are \textbf{information-seeking} in that they are written without seeing the answer, as is the case with real users of question answering systems. We contrast this with the reading comprehension task where the question-writer sees the answer passage prior to writing the question; this genre of questions tends to have both higher lexical overlap with the question and elicit questions that may not be of broad interest.} are paired with retrieved documents in which annotators identify an answer if one is available \cite{kwiatkowski2019natural}. As many African languages lack high-quality in-language content online, \afriqa employs a cross-lingual setting \cite{asai-etal-2021-xor} where relevant passages are retrieved in a high-resource language spoken in the corresponding region and answers are translated into the source language.
To ensure utility of this dataset, we carefully select a relevant source language (either English or French) based on its prevalence in the region corresponding to the query language.
\afriqa includes 12,000+ examples across 10 languages spoken in different parts of Africa.
The majority of the dataset's questions are centered around entities and topics that are closely linked to Africa.
This is an advantage over simply translating existing datasets into these languages. 
By building a dataset from the ground up that is specifically tailored to African languages and their corresponding cultures, we are able to ensure better contextual relevance and usefulness of this dataset. 

\insertSampledData

We conduct baseline experiments for each part of the open-retrieval QA pipeline using different translation systems, retrieval models, and multilingual reader models. We demonstrate that cross-lingual retrieval still has a large deficit compared to automatic translation and retrieval; we also show that a hybrid approach of sparse and dense retrieval improves over either technique in isolation. We highlight interesting aspects of the data and discuss annotation challenges that may inform future annotation efforts for QA. Overall, \afriqa proves challenging for state-of-the-art QA models. We hope that \afriqa encourages and enables the development and evaluation of more multilingual and equitable QA technology.
The dataset is released under the Creative Commons Attribution 4.0 International (CC BY 4.0) license\footnote{\url{https://creativecommons.org/licenses/by/4.0/}}.

In summary, we make the following contributions:
\begin{itemize}
    \item We introduce the first cross-lingual question answering dataset with 12,000+ questions across 10 geographically diverse African languages. This dataset directly addresses the deficit of African languages in existing datasets.

    \item We conduct a comprehensive analysis of the linguistic properties of the 10 languages, which is crucial to take into account when formulating questions in these languages.
    
    \item Finally, we conduct a comprehensive evaluation of the dataset for each part of the open-retrieval QA pipeline using various translation systems, retrieval models, and multilingual reader models.
\end{itemize}


\section{\afriqa}
The \afriqa dataset was created by researchers from Masakhane\footnote{\url{https://www.masakhane.io/}}---a not-for-profit community that promotes the representation and coverage of under-resourced African languages in NLP research---in collaboration with Google. We show examples of the data in Table \ref{tab:dataset_examples}. In \textsection\ref{sec:language_discussion}, we provide an overview of the 10 languages discussing their linguistic properties, while \textsection\ref{data-collection} and \textsection\ref{quality-control}  describe the data collection procedure and quality control measures put in place to ensure the quality of the dataset.

\subsection{Discussion of Languages}
\label{sec:language_discussion}

African languages have unique typologies, grammatical structures, and phonology, many of them being tonal and morphologically rich \cite{adelani-etal-2022-masakhaner}.
We provide an overview of the linguistic properties of the ten languages in \afriqa that are essential to consider when crafting questions for QA systems.

\smallskip
\noindent
\textbf{Bemba} is a morphologically rich language like many Bantu languages, which attaches affixes to the headword to change grammatical forms such as tenses, negation, and plurality when formulating questions.
Negation is typically expressed using three morphemes: ``ta-'' (e.g. \textbf{ta}baleelanda -- They are not speaking), ``-shi-'' (e.g. aba\textbf{shi}leelanda  -- who is not talking ), and ``kaana'' (e.g. uku\textbf{kaana} lya  -- not eating).
The present tense is typically indicated by ``ali'' (e.g. Ninaani \textbf{ali} kateeka wa caalo ca Zambia? -- Who is the president of Zambia?
) and past tense by ``aali'' (e.g. Ninaani \textbf{aali} kateeka wa caalo ca Zambia?  -- Who was the former president of Zambia?
).
Plurality is indicated by prefixes attached to the stem of the noun, which vary according to the noun class they belong to. For example, ``u-\textbf{mu}-ntu'' (person), and ``a-\textbf{ba}-ntu'' (people).
Typical question wh-words used are ``cinshi'' (what), ``naani''(who), ``liisa'' (when), ``mulandunshi'' (why), ``ciisa'' (which), ``kwi/kwiisa'' (where), and shaani (how).

\smallskip
\noindent
\textbf{Fon} is an isolating language in terms of morphology typology. For changes in grammatical forms such as tenses, negation, and plurality when formulating questions, a new word is added to express this change. For example, ``ǎ'' is added for negation, ``xóxó'' to indicate past tense, and ``l\'\textepsilon ''for plurality. Common question wh-words are Et\'\textepsilon (what), M\textepsilon \ (who),  Hwet\'\textepsilon nu (when), Aniw\'{u} (why), \textrtaild e t\textepsilon \ (which) and Fit\textepsilon \ (where).

\smallskip
\noindent
\textbf{Hausa} is the only Afro-Asiatic language in \afriqa.
It typically makes use of indicative words for changes to the grammatical forms within a sentence, such as negation, tenses, and plurality. 
For negation, the indicative words ``ba/ba a'' (not) and ``banda'' (except) are used.
For tenses, ``tsohon'' (was) and ``yanzu'' (is) are used to indicate past and present tenses. 
Plurality has complex forms which often require the deletion of the last vowel of the word and the addition of a suffix (like ``una'', ``aye'' and ``oli''). For example,  ``hula'' (cap) -- ``huluna'' (caps),  ``mace'' (girl) -- ``mataye'' (girls). Typical question wh-words used are ``me/ya'' (what), ``wa''(who), ``yaushe'' (when), ``dan me/akan me'' (why), ``wanne'' (which), ``ina/ a ina'' (where), and ``yaya/qaqa'' (how). 

\smallskip
\noindent
\textbf{Igbo} is a morphologically rich language and most changes in grammatical forms (negations, questions) can be embedded in a single word or by varying the tone.
For example, a suffix ``ghi'' often signifies a negation.
However, there is no affix to indicate plurality, the count is often specified after the word. Question words are often preceded by ``kedu'' or ``gini'' like ``kedu/gini'' (what), ``onye/kedu onye'' (who), ``kedu mgbe'' (when), ``gini mere/gini kpatara'' (why), ``kedu nke'' (which), ``ebee'' (where), and ``kedu ka'' or ``kedu etu'' (how).

\smallskip
\noindent
\textbf{Kinyarwanda} is a morphologically rich language with several grammatical features such as negation, tenses, and plurals that are expressed as changes to morphemes in a word. For example, the plural of ``u\textbf{mu}ntu'' (person) is  ``\textbf{aba}ntu'' (people).
According to \cite{jarnow2020endless}, Kinyarwanda lacks an overt question particle or a syntactic movement process to form polar questions (yes/no).
Thus, Kinyarwanda makes use of prosodic and tonological processes to differentiate between declarative and polar questions. 
Question words typically used are ``iki'' (what), ``nde/inde'' (who), `ryari`'' (when), ``ikihe/uwuhe'' (which), ``hehe'' (where), and ``gute'' (how).

\smallskip
\noindent
\textbf{Swahili} is a morphologically rich language that typically has several morphemes to incorporate changes to grammatical forms such as negation, tenses and plurality. For example, ``ni'' (is), ``alikuwa'' (was/former), and ``atakuwa'' (will be) indicate present, past, and future tenses.
Similar to Kinyarwanda, changes to the prefix indicate plurality, for example, ``\textbf{m}tu'' (person) --- ``\textbf{wa}tu'' (people), ``gari'' (car) --- ``\textbf{ma}gari''(cars).
A question word can be placed at the beginning or end of the question sentence, for example, ``amekuja nani?'' (who has come?) and ``nani amekuja?'' (who has come?).
The question word ``gani'' requires a noun modifier and can be placed at the beginning or end of the sentence. Other question words often used are ``nini'' (what), ``nani'' (who), ``lini'' (when), ``kwanini'' (why), ``wapi'', (where), and ``vipi'' (how).

\smallskip
\noindent
\textbf{Twi} is a dialect of the Akan language and \afriqa includes the Asante variant. Akan makes extensive use of different affixes for different grammatical changes such as negation, plurality, and tenses. For example, the suffix ``n'' is added to the root word to express negation. Similarly, plurality is indicated by replacing the first two letters of a root word with ``mm'' or ``nn'', and in some cases, a suffix ``nom'' can be used instead of a prefix. A few common question words used are ``\textepsilon de\textepsilon n''(what), ``hwan'' (who), ``daben'' (when), ``adεn'' (why), ``de\textepsilon hen'' (which), ``\textepsilon henfa'' (where), ``s\textepsilon n'' (how).

\smallskip
\noindent
\textbf{Wolof} is an agglutinative language, however, it does not use an affix attached to the headwords like in Bantu languages. Instead, it makes use of a dependent such as a determiner that is not attached to the headword. Changes to the grammatical form like negation, tenses, and plurality are captured by this dependent word. For incorporating past tense, ``oon'' is attached to the end of the verb, while for plurality, ``yi'' or ``ay'' is attached before or after the word. For example, ``xar mi''(a sheep) -- ``xar yi''(the sheeps).
For negation, suffix ``ul'' is added at the end of the verb, for example ``man réew moo \textbf{nekk} ci Afrig'' (which one of these countries is \textbf{located} in Africa?) --- ``man réew moo \textbf{nekkul} ci Afrig.'' (which one of these countries is \textbf{not located} in Africa?). A few common question words used are ``ian''(what), ``kan'' (who), ``kañ'' (when), ``lu tax'', ``ban'' (which), ``fan'' (where), and ``naka'' (how). 

\smallskip
\noindent
\textbf{\yoruba} has a derivational morphology that entails affixation, reduplication, and compounding.
However, there is no affix to indicate plurality in the language; a number is instead specified with a separate token.
\yoruba employs polar question words such as ``nje'', ``se'', ``abi'', ``sebi'' (for English question words ``do'' or ``is'', ``are'', ``was'' or ``were'') and content question markers such as ``tani'' (who), ``kini'' (what), ``nibo'' (where), ``elo/meloo'' (how many), ``bawo''(how is), ``kilode'' (why), and ``igba/nigba'' (when).
Negation can be expressed with ``k\`{o}''. 
Phonological rules must be followed when adapting a loanword to \yoruba.
For instance, if the loanword has consonant clusters, a vowel might be added in between the clusters or the phonological structures of the clusters might be modified.

\smallskip
\noindent
\textbf{Zulu} is a very morphologically-rich language where several grammatical features such as tense, negation, and the plurality of words are indicated through prefixes or suffixes. Negation is typically indicated by the prefix ``nga-''.
The present tense is indicated by an affix after the subject concord.
For example, ``ya'' or ``sa'' indicates present tense (as in ``\textit{Ngi\textbf{ya}dlala}'' -- I am playing), while past tense is indicated by a suffix, for example ``e'' or ``ile'' (in ``\textit{Ngikhathal\textbf{ile}}'') -- I was tired). The most commonly used question words are ``yini'' (what), ``ubani'' (who), ``nini'' (when), ``kungani'' (why), ``yiliphi'' (which), ``kuphi'' (where), ``kanjani'' (how), and ``yenza'' (do).

\subsection{Data Collection Procedure}
\label{data-collection}

For each of the 10 languages in \afriqa, a team of 2--6 native speakers was responsible for the data collection and annotation. Each team was led by a coordinator.
The annotation pipeline consisted of 4 distinct stages: 1) question elicitation in an African language; 2) translation of questions into a pivot language; 3) answer labeling in the pivot language based on a set of candidate paragraphs; and 4) answer translation back to the source language. All data contributions were compensated financially.

\subsubsection{Question Elicitation}
The TyDi QA methodology \cite{clark-etal-2020-tydi} was followed to elicit locally relevant questions.
Team members were presented with prompts including the first 250 characters of the most popular Wikipedia\footnote{\url{https://www.wikipedia.org/}} articles in their languages, and asked to write factual or procedural questions for which the answers were not contained in the prompts. Annotators were encouraged to follow their natural curiosity. 
This annotation process avoids excessive and artificial overlap between the question and answer passage, which can often arise in data collection efforts for non-information-seeking QA tasks such as reading comprehension.\footnote{Reading comprehension differs from information-seeking QA as question-writers see the answer prior to writing the question and thus tests understanding of the answer text rather than the general ability to provide a correct answer.} 
For Fon and Bemba where there is no in-language Wikipedia, team members were presented with prompts relevant to Benin and Zambia from the French and English Wikipedia respectively, and asked to generate questions in their native languages.
For Swahili, questions elicited in TyDi QA, which remained unanswered in the original dataset were used with light curation from the Swahili team for correctness.
These questions remained unanswered because the TyDi QA annotator team was not able to find a candidate paragraph in Swahili to answer them. The question elicitation was carried out via simple spreadsheets.

Before moving on to the second stage, team coordinators reviewed elicited questions for grammatical correctness and suitability for the purposes of information-seeking QA.\footnote{For example, personal questions such as \textit{How are you?} and opinion questions such as \textit{What is the best dessert?} were excluded.}

\subsubsection{Question Translation}
Elicited questions were translated from the original African languages into pivot languages following \citet{asai-etal-2021-xor}.
English was used as the pivot language across all languages except Wolof and Fon, for which French was used.\footnote{French is widely used in the regions where Fon and Wolof are spoken, so there may be a higher probability of finding answers in French than in other pivot languages.}
Where possible, questions elicited by one team member were allocated to a different team member for translation to further ensure that only factual or procedural questions that are grammatically correct make it into the final dataset. This serves as an additional validation layer for the elicited questions.

\subsubsection{Answer Retrieval}
Using the translated questions as queries, Google Programmable Search Engine\footnote{\url{https://developers.google.com/custom-search/}} was used to retrieve Wikipedia paragraphs that are candidates to contain an answer in the corresponding pivot language.
The Mechanical Turk interface\footnote{The Mechanical Turk \textit{interface} was used, but no Mechanical Turk \textit{workers} were employed---all annotations were carried out by team members.} was used to show candidate paragraphs to team members who were then asked to identify 1) the paragraph that contains an answer and 2) the exact minimal span of the answer.
In the case of polar questions, team members had to select ``Yes'' or ``No'' instead of the minimal span. 
In cases where  candidate paragraphs did not contain the answer to the corresponding question, team members were instructed to select the ``No gold paragraph'' option.

As with question elicitation, team members went through a phase of training, which included a group meeting where guidelines were shared and annotators were walked through the labeling tool. 
Two rounds of in-tool labeling training were conducted.

\insertdatasetstatistics

\subsubsection{Answer Translation}
To obtain answers in the African languages, we translated the answers in the pivot languages to the corresponding African languages. We allocated the task of translating the answers labeled by team members to different team members in order to ensure accuracy. Translators were instructed to minimize the span of the translated answers.
In cases where the selected answers were incorrect or annotators failed to select the minimum span, we either removed the question, corrected the answer, or re-annotated the question using the annotation tool.

\subsection{Quality Control}
\label{quality-control}

To ensure completeness, quality, and suitability of the dataset, we implemented rigorous quality control measures at every stage of the dataset creation process. We recruited only native speakers of the languages as annotators and team coordinators. Prior to eliciting questions in their native languages, annotators underwent three rounds of training in question elicitation using English prompts. Each annotator received personalized feedback during each training round, with a focus on ensuring that the elicited questions were factual and that the answers were not present in the prompts. Only annotators that achieved a minimum accuracy rate of 90\% were permitted to proceed with the question elicitation in their native languages. For annotators who were unable to achieve the target percentage, additional training rounds with one-on-one instruction were provided. Both annotators and team coordinators participated in the question elicitation training.

All language teams consisted of at least 3 members, including a coordinator, except for Fon and Kinyarwanda teams, which had 2 members. This was done to ensure that the questions elicited by one team member were translated by another team member for quality control purposes. During the question translation phase, annotators were asked to flag questions that were not factual. These questions were either corrected or removed from the datasets. Similarly, during the answer labeling phase, annotators were provided with comment options to indicate if a question was unsuitable for the datasets, which were then used to filter out questions. Furthermore, language team coordinators reviewed the question-and-answer pairs alongside their translations, while central project managers reviewed the translations for consistency. Common issues were identified, such as answer-span length, accidental selection of Yes/No when the question is not polar or vice versa, and wrong answer selection. Span lengths were fixed in post-production, while wrong answers or polar question misunderstandings resulted in questions being removed from the dataset.

\subsection{Final Dataset}
The statistics of the dataset are presented in \autoref{tab:datastats}, which includes information on the languages, their corresponding pivot languages, and the total number of questions collected for each language.
The final dataset consists of a total of 12,239 questions across 10 different languages, with 8,892 corresponding question-answer pairs. We observed a high answer coverage rate, with only 27\% of the total questions being unanswerable. This can be attributed to the lack of relevant information on Wikipedia, especially for named entities with sparse information.
Despite this sparsity, we were able to find answers for over 60\% of the questions in most of the languages in our collection.

\section{Tasks and Baselines}

As part of our evaluation for \afriqa, we follow the methodology proposed in \citet{asai-etal-2021-xor} and assess its performance on three different tasks: \texttt{XOR-Retrieve}, \texttt{XOR-PivotLanguageSpan}, and \texttt{XOR-Full}.
Each task poses unique challenges for cross-lingual information retrieval and question answering due to the low-resource nature of many African languages.

\subsection{XOR-Retrieve}
The XOR-Retrieve task focuses on cross-lingual passage retrieval. 
Specifically, given a question $q_x$ in language $X$, the goal is to find a set of passages in a pivot language $Y$ that contains an answer to the question.
This task is particularly challenging for African languages due to the limited availability of resources, which makes it difficult to retrieve relevant passages in the source language or pivot language.
For our experiments, we measure the retrieval effectiveness using recall@$k$, as defined in \citet{karpukhin-etal-2020-dense}, where $k \in {10, 20, 100}$. The recall@$k$ is calculated as the percentage of questions for which the answer span appears in one of the top $k$ retrieved passages.

\smallskip
\noindent
\textbf{Retrieval Corpora:}
We use Wikipedia as the retrieval corpus for the XOR experiments.
Specifically, we use processed Wikipedia dumps in English and French as our retrieval passage corpora, as these are our pivot languages. 
More information on the processing details can be found in \autoref{sec-prepwiki}.

\subsection{XOR-PivotLanguageSpan}

This task is designed to address the challenge of answering questions in language $X$, using passages in a pivot language $Y$. Specifically, given a question $q_x$ in language $X$, the goal is to identify a set of passages in language $Y$ that contain the answer to $q_x$ and extract the answer span $a_y$ from these passages. We also include baselines for extracting the answer span from annotated gold passages for that question.
We evaluate the effectiveness of our predictions using the Exact Match (EM) accuracy and F1 metrics, as outlined in \citet{2016arXiv160605250R}.
This evaluation is based on how much the predicted answer spans match the token set of the correct answer.

\subsection{XOR-Full}
This task is similar to \texttt{XOR-PivotLanguageSpan}, with the difference being that we are trying to find answers to a question in the same language as the question.
Specifically, given a question $q_x$ in language $X$, the goal is to find an answer span $a_x$ in the same language while leveraging passages in a pivot language $Y$ and translating the answer back to the question language.
We evaluate this task using the same metrics (F1 and EM) as the \texttt{XOR-PivotLanguageSpan} task. In addition, we also include BLEU scores to measure the degree of overlap between translated answer spans and ground-truth human translations.

\section{Experiments}

In this section, we describe the different baseline translation, retrieval, and reading comprehension systems.

\subsection{Translation Systems}
\label{sec:translation_system}

A common approach to cross-lingual question answering is to translate queries from the source language into a target language, which is then used to find an answer in a given passage. This approach requires the use of translation systems that can accurately translate the queries from one language to another.
For our experiments, we explore the use of different translation systems as baselines for \afriqa.
We consider human translation, Google Translate, and open-source translation models such as NLLB \cite{https://doi.org/10.48550/arxiv.2207.04672} and fine-tuned M2M-100 models \cite{adelani-etal-2022-thousand} in zero-shot settings.
Below is a breakdown of the different machine translation systems.

\paragraph{Google Machine Translation.} 
We use Google Translate because it is readily available and provides out-of-the-box translation for 7 out of 10 languages in our dataset.
Although Google Translate provides a strong translation baseline for many of the languages, we cannot guarantee the future reproducibility of these translations as it is a product API and is constantly being updated. 
For our experiments, we use the translation system as of February 2023. Note that while Google Translate supports 133 languages, it does not include Bemba, Fon, nor Wolof; this speaks to the very low-resource nature of the languages included in this work and the difficulty of building systems for them.

\paragraph{NLLB.}
NLLB is an open-source translation system trained on 100+ languages and provides translation for all the languages in \afriqa.
At the time of release, NLLB provides state--of--the--art translation in many languages and covers all the languages in our dataset.
For our experiments, we use the 1.3B size NLLB models.\footnote{\url{https://huggingface.co/facebook/nllb-200-1.3B}}

\paragraph{MAFAND M2M-100.}
MAFAND M2M-100 is an adaptation of the M2M-100~\cite{10.5555/3546258.3546365} machine translation model to 16 African languages in the news domain~\citep{adelani-etal-2022-thousand}. Each translation direction (e.g., yor--eng) was fine-tuned on a few thousand (2.5k--30K) parallel sentences in the news domain.

\insertMTblue

\smallskip
\noindent
\autoref{tab:mt_bleu} shows the BLEU score of the different translation systems on the test set of \afriqa, evaluated against the human-translated queries. Google Translate performs the best on the languages it supports while NLLB 1.3B achieves slightly poorer performance with a broader language coverage.

\subsection{Passage Retrieval}
\label{subsec-passageretrieval}

We present two baseline retrieval systems: translate--retrieve and cross-lingual baselines.
In the translate--retrieve baseline, we first translate the queries using the translation systems described in  \textsection\ref{sec:translation_system}. The translated queries are used to retrieve relevant passages using three different retrieval systems: BM25, multilingual Dense Passage Retriever (mDPR), and a hybrid combination of BM25 and mDPR. 
Alternatively, the cross-lingual baseline directly retrieves passages in the pivot language without the need for translation using a multilingual dense retriever.

\smallskip
\noindent
\textbf{BM25.} 
BM25 \cite{bm25} is a classic term-frequency-based retrieval model that matches queries to relevant passages using the frequency of word occurrences in both queries and passages.
 We use the BM25 implementation provided by Pyserini \cite{Lin_etal_SIGIR2021_Pyserini} with default hyperparameters k1 = 0.9, b = 0.4 for all languages.

\smallskip
\noindent
\textbf{mDPR.} We evaluate the performance of mDPR, a multilingual adaptation of the Dense Passage Retriever (DPR) model \citep{karpukhin-etal-2020-dense}.
In mDPR, we replace the BERT model in DPR with multilingual BERT (mBERT) which is fine-tuned on the MS MARCO passage ranking dataset~\cite{msmarco}.
While this approach has been found effective for monolingual retrieval \cite{Zhang_etal_arXiv2022}, we also investigate its potential for cross-lingual retrieval by using original language queries for passage retrieval and translated queries for monolingual retrieval. Retrieval is performed using the Faiss flat index implementation provided by Pyserini.

\smallskip
\noindent
\textbf{Sparse--Dense Hybrid.}
We also explore sparse-dense hybrid baselines, a combination of sparse (BM25) and hybrid (mDPR) retrievers.
We use a linear combination of both systems to generate a reranked list of passages for each question.

\subsection{Answer Span Prediction}
\label{subsec-answer-span-prediction}

To benchmark models' answer selection capabilities on \afriqa, we combine different translation, extractive, and generative QA approaches.

\smallskip
\noindent
\textbf{Extractive QA on Gold Passages.}
In this approach, we extract the answer span from passages that have been manually annotated in both French and English, using both original and translated queries. We used AfroXLMR~\cite{alabi2022adapting} as a backbone to train our extractive QA models. The models were trained on SQuAD 2.0~\cite{2016arXiv160605250R} and FQuAD \cite{dhoffschmidt-etal-2020-fquad} separately.

\smallskip
\noindent
\textbf{Generative QA on Gold Passages.}
To evaluate the performance of generative question answering, we utilize mT5--base~\cite{xue-etal-2021-mt5} fine-tuned on SQuAD 2.0~\cite{2016arXiv160605250R} and evaluate it using both translated and original queries.
The model was provided with the queries and the gold passages that were annotated using a template prompt and generates the answers to the questions.

\smallskip
\noindent
\textbf{Extractive QA on Retrieved Passages.}
For \texttt{XOR-PivotLanguageSpan} baselines, we employed an extractive question-answering model that extracts the answer span from the output passages produced by the various retrieval baselines outlined in \textsection\ref{subsec-passageretrieval}.
The model is trained to extract answer spans from each passage, along with the probability indicating the likelihood of each answer.
The answer span with the highest probability is selected as the correct answer. We trained a multilingual DPR reader model, which was initialized from mBERT and trained on Natural Questions \cite{kwiatkowski2019natural}.
\section{Results and Analysis}

\insertretrievalexperimentstopten

\insertpivotlangspanmt


\subsection{XOR-Retrieve Results}
\begin{figure}[t]
\includegraphics[trim=0 0em 0 0,clip,width=\columnwidth]{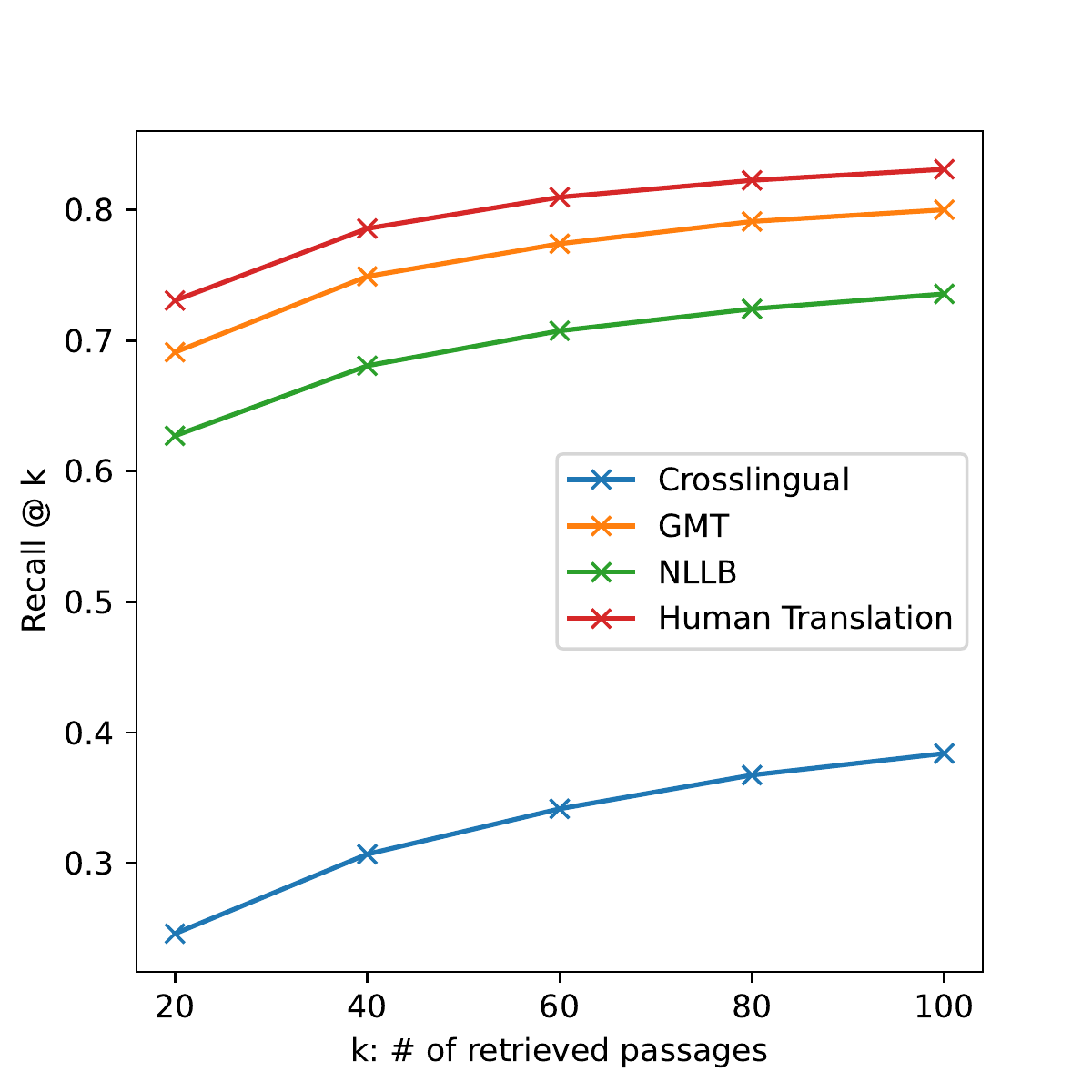}
\vspace{-2em}
\caption{Graph of retriever recall@k for different translation systems. The scores shown in this graph are from mDPR retrieval.}
\label{fig:topk}
\vspace{-1em}
\end{figure}

We present the retrieval results for recall@10 and recall@100 in \autoref{tab:retrievalexperimentstopten}.\footnote{For recall@k retrieval results, we assume that there is only one gold passage despite the possibility of other retrieved passages containing the answer.}
The table includes retriever results using different question translations and retrieval systems.
We also report the performance with both original and human-translated queries.
The table shows that hybrid retrieval using human translation yields the best results for all languages, with an average recall@10 of 73.9 and recall@100 of 86.2. 
In isolation, mDPR retrieval outperforms BM25 for all translation types.
This table also enables us to compare the effectiveness of different translation systems in locating relevant passages for cross-lingual question answering in African languages.
This is illustrated in \autoref{fig:topk}, showing retriever recall rates for different translation types at various cutoffs using mDPR.

We observe that human translation yields better accuracy than all other translation types, indicating that the current state-of-the-art machine translation systems still have a long way to go in accurately translating African languages. Google Translate shows better results for the languages where it is available, while the NLLB model provides better coverage.  
The cross-lingual retrieval model that retrieves passages using questions in their original language is the least effective of all the model types. 
This illustrates that the cross-lingual representations learned by current retrieval methods are not yet of sufficient quality to enable accurate retrieval across different languages.

\insertpivotlangspan

\subsection{XOR-PivotLanguageSpan Results}

\smallskip
\noindent \label{extractive-qa-results}\textbf{Gold Passage Answer Prediction.} We first evaluate the extractive and generative QA setting using gold passages.
We present F1 and Exact Match results using different methods to translate the query in \autoref{tab:pivotlangspanmt5} and \autoref{tab:pivotlangspanextractive}.
On both approaches, human translation of the  queries consistently outperforms using machine-translated queries, which outperforms using queries in their original language.
The generative setting using mT5 yields slightly better results on average compared to the extractive setting  across different translation systems.

\insertpivotlanguagespan

\smallskip
\noindent
\textbf{Retrieved Passages Answer Prediction.} 
We now evaluate performance using retrieved passages. We present F1 and Exact Match results with different translation--retriever combinations in \autoref{tab:pivot_lang_span_retrieve}.
We extract the answer spans from only the top-10 retrieved passages for each question using an extractive multilingual reader model (see \textsection\ref{subsec-answer-span-prediction}). The model assigns a probability to each answer span, and we select the answer with the highest probability as the final answer.


Our results show that hybrid retrieval using human-translated queries achieves the best performance across all languages on average.  
Using human-translated queries generally outperforms using translations by both Google Translate and NLLB, regardless of the retriever system used.
In terms of retrieval methods, mDPR generally performs better than BM25, with an average gain of 3 F1 points across different translation types.
These results highlight the importance of carefully selecting translation--retriever combinations to achieve the best answer span prediction results in cross-lingual question answering.


\subsection{XOR-Full Results}

Each pipeline consists of components for question translation, passage retrieval, answer extraction, and answer translation.
From \autoref{tab:fulltrip}, we observe that Google machine translation combined with mDPR is the most effective.
This is followed by a pipeline combining NLLB translation with mDPR.

\insertxorfull

\section{Related Work}
\smallskip
\noindent \textbf{Africa NLP.}
In parallel with efforts to include more low-resource languages in NLP research~\cite{costa2022no,ruder2020you}, demand for NLP that targets African languages, which represent more than 30\% of the world’s spoken languages~\cite{ogueji2022afriberta} is growing. This has resulted in the creation of publicly available multilingual datasets targeting African languages for a variety of NLP tasks such as sentiment analysis~\cite{muhammad2023afrisenti, shode2022yosm}, language identification~\cite{adebara2022afrolid}, data-to-text generation~\cite{gehrmann2022tata}, topic classification~\cite{adelani2023masakhanews,hedderich2020transfer}, machine translation~\cite{adelani-etal-2022-thousand,nekoto2020participatory}, and NER~\cite{eiselen2016government,adelani-etal-2021-masakhaner,adelani-etal-2022-masakhaner}.

Datasets for QA and Information Retrieval tasks have also been created. They are, however, very few and cater to individual languages~\cite{abedissa2023amqa,10.1145/3578553} or a small subset of languages spoken in individual countries~\cite{daniel2019towards, https://doi.org/10.48550/arxiv.2210.09984}. 
Given the region’s large number of linguistically diverse and information-scarce languages, multilingual and cross-lingual datasets are encouraged to catalyze research efforts. To the best of our knowledge, there are no publicly available cross-lingual open-retrieval African language QA datasets.

\smallskip
\noindent \textbf{Comparison to Other Resources.}
Multilingual QA datasets have paved the way for language models to simultaneously learn across multiple languages, with both reading comprehension~\cite{lewis-etal-2020-mlqa} and other QA datasets~\cite{longpre-etal-2021-mkqa,clark-etal-2020-tydi} predominantly utilizing publicly available data sources such as Wikipedia, SQUAD, and the Natural Questions dataset. To address the information scarcity of the typically used data sources for low-resource languages, cross-lingual datasets~\cite{liu-etal-2019-xqa,asai-etal-2021-xor} emerged that translate between low-resource and high-resource languages, thus providing access to a larger information retrieval pool which decreases the fraction of unanswerable questions. Despite these efforts, however, the inclusion of African languages remains extremely rare, as shown in ~\autoref{tab:corpuscomparison}, which compares our dataset to other closely related QA datasets. TyDi QA features Swahili as the sole African language out of the 11 languages it covers.

In recent years, efforts to create cross-lingual information retrieval datasets that include African languages have resulted in the creation of datasets such as AfriCLIRMatrix~\cite{ogundepo-etal-2022-africlirmatrix} and CLIRMatrix~\cite{sun2020clirmatrix} which feature 15 and 5 African languages respectively. These CLIR datasets however are not specific to QA and are synthetically generated from Wikipedia.

\section{Conclusion}

In this work, we take a step toward bridging the information gap between native speakers of many African languages and the vast amount of digital information available on the web by creating \afriqa, the first cross-lingual question-answering dataset focused on African languages.
\afriqa is an open-retrieval question answering with 12,000+ questions across 10 African languages.
We evaluate our dataset on cross-lingual retrieval and reading comprehension tasks.

We anticipate that \afriqa will help improve access to relevant information for speakers of African languages. 
By leveraging the power of cross-lingual question answering, we hope to bridge the information gap and promote linguistic diversity and inclusivity in digital information access. 
Overall, this work represents a crucial step towards democratizing access to information and empowering underrepresented African communities by providing tools to engage with digital content in their native languages.

\section*{Acknowledgements}
We would like to thank Google Cloud for providing us access to computational resources through free cloud credits.
We are grateful to Google Research for funding the dataset creation.
Finally, we thank Knowledge4All for their administrative support throughout the project.

\section*{Contributions}
\label{sec:contributions}

In this section, we provide more details about the contributions of each author.

{\small
\textbf{Data Annotation}: Andre  Niyongabo Rubungo, Boyd  Sinkala, Daniel Abidemi Ajisafe, Emeka Felix  Onwuegbuzia, Emile  Niyomutabazi, Eunice  Mukonde, Falalu Ibrahim LAWAN, Habib  MBOW, Ibrahim Said Ahmad, Jesujoba O. Alabi, Martin  Namukombo, Mbonu Chinedu Emmanuel, Mofetoluwa  Adeyemi, Mofya  Phiri, Ndumiso  Mngoma, Neo  Putini, Orevaoghene  Ahia, Priscilla  Amondi  Amuok, Ruqayya Nasir Iro,Sonia  Adhiambo, Albert Njoroge Kahira, Aremu Anuoluwapo, Ayodele  Awokoya, Bernard Opoku, Chiamaka Chukwuneke, Christine  Mwase, Clemencia Siro, Oyinkansola Fiyinfoluwa Awosan, Steven Arthur, Shamsuddeen Hassan Muhammad, Tunde Oluwaseyi Ajayi, Verrah Otiende, Chris Emezue, Claytone Sikasote, David Adelani, Happy Buzaaba, Ignatius Ezeani, Rooweither Mabuya, Salomey Osei, Abdou Aziz DIOP, Bonaventure F. P. Dossou, Gilles Hacheme\\
\textbf{Language Team Coordination}: Chris Emezue, Claytone Sikasote, David Adelani, Happy Buzaaba, Ignatius Ezeani, Rooweither Mabuya, Salomey Osei, Abdou Aziz DIOP, Albert Njoroge Kahira, Shamsuddeen Hassan Muhammad, Bonaventure F. P. Dossou, Gilles Hacheme\\
\textbf{Paper Writing}:  Ogundepo Odunayo, David Adelani, Jonathan H. Clark, Sebastian Ruder, Clara E. Rivera,  Tajuddeen R. Gwadabe, Tunde Oluwaseyi Ajayi, Chris Emezue, Claytone Sikasote, Happy Buzaaba, Ignatius Ezeani, Rooweither Mabuya, Salomey Osei, Abdou Aziz Diop, Abraham Toluwase Owodunni, Atnafu Lambebo Tonja, Iyanuoluwa Shode, Bernard Opoku, Chiamaka Chukwuneke, Christine  Mwase, Clemencia Siro, Aremu Anuoluwapo, Ayodele  Awokoya, Oyinkansola Fiyinfoluwa AWOSAN, Steven Arthur, Verrah Otiende\\
\textbf{Output Data Preparation and Experimentation}: Ogundepo Odunayo, Abraham Toluwase Owodunni, Atnafu Lambebo Tonja, Iyanuoluwa Shode, Abdou Aziz DIOP\\
\textbf{Annotator Training}: Jonathan H. Clark, Sebastian Ruder, Clara E. Rivera, Tajuddeen R. Gwadabe\\
\textbf{Task Framing}: Jonathan H. Clark, Sebastian Ruder, Clara E. Rivera, Tajuddeen R. Gwadabe, Akari Asai, Ogundepo Odunayo\\
\textbf{Project Coordination}: Clara E. Rivera, Tajuddeen R. Gwadabe, Ogundepo Odunayo\\
\textbf{Documentation}: Clara E. Rivera, Tajuddeen R. Gwadabe, Ogundepo Odunayo\\
\textbf{Input Data Preparation and Annotation Tool Management}: Bonaventure F. P. Dossou, Gilles Hacheme\\
\textbf{Quality Control}: Clara E. Rivera Tajuddeen R. Gwadabe, Ogundepo Odunayo, Jonathan H. Clark, Sebastian Ruder
}

\bibliography{main}
\bibliographystyle{acl_natbib}

\clearpage
\section{Appendix}
\appendix

\section{Preparing Wikipedia Passages}
\label{sec-prepwiki}

\inserttrainingparams

\insertxorfullem

\begin{figure*}[t]
    \centering
    \vspace{-0.3cm}
    \includegraphics[width=\textwidth]{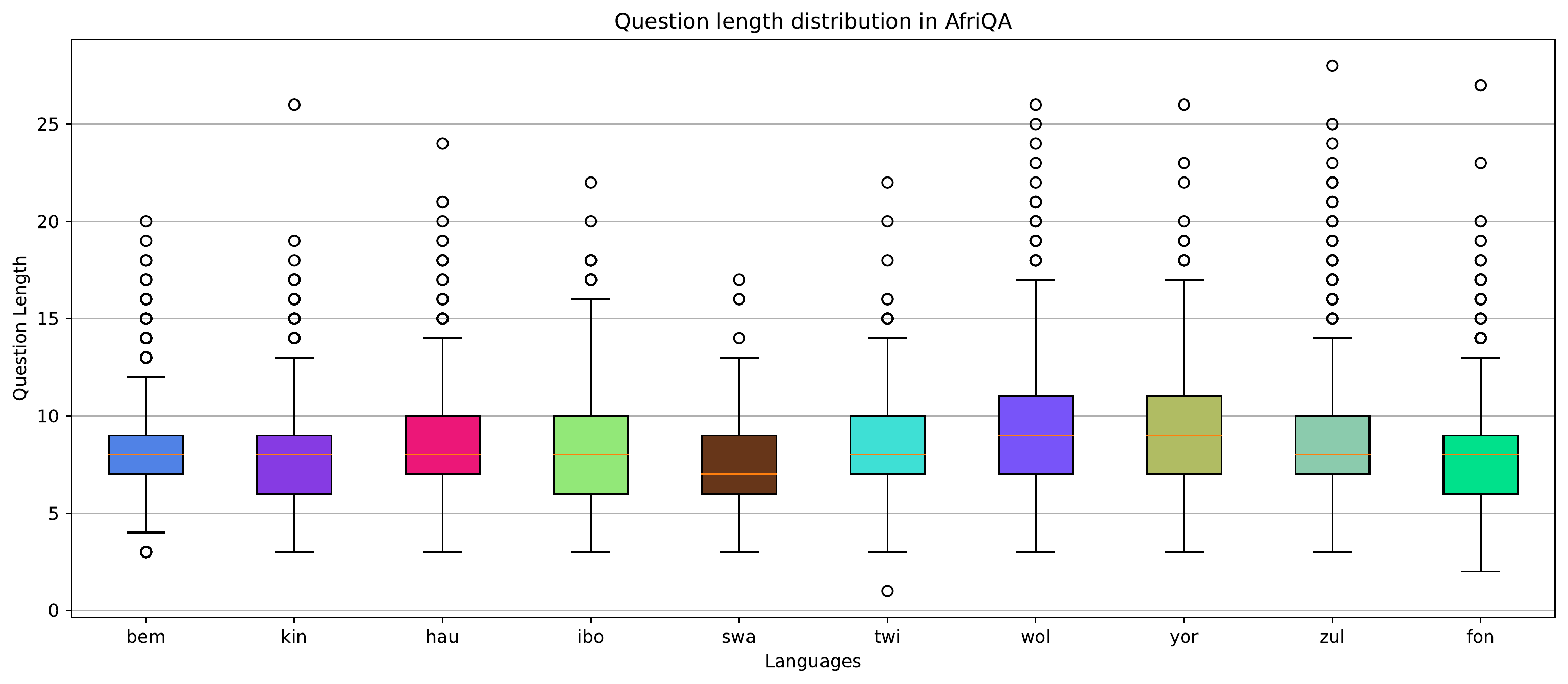}
    \vspace{-0.75cm}
    \caption{Question Length Distribution}
    \label{fig:question_length}
\end{figure*}

Wikipedia is a popular choice as a knowledge base for open-retrieval question-answering (QA) experiments, where articles are usually divided into fixed-length passages that are indexed and used for retrieval and reading comprehension, as seen in previous works such as \cite{karpukhin-etal-2020-dense, asai-etal-2021-xor}.
However, \citet{manveer_preprocessing} highlighted that splitting articles into fragmented and disjoint passages can negatively impact downstream reading comprehension performance. Instead, they proposed a sliding window segmentation approach to create passages from Wikipedia articles. In line with this methodology, we used the same approach to create passages for our cross-lingual question-answering experiments.

To create our passages, we downloaded the Wikipedia dumps dated May 01, 2022, for English Wikipedia \footnote{\url{https://archive.org/download/enwiki-20220501/enwiki-20220501-pages-articles-multistream.xml.bz2}} and April 20, 2022, for French Wikipedia \footnote{\url{https://archive.org/download/frwiki-20220420/frwiki-20220420-pages-articles-multistream.xml.bz2}}. We then applied the sliding window approach to generate fixed-length passages of 100 tokens each from these dumps. These passages serve as our knowledge base for retrieval and answer span extraction.
By adopting the sliding window segmentation approach for creating Wikipedia passages, we aim to improve downstream reading comprehension performance. The fixed-length passages enable efficient indexing and retrieval of relevant information for a given question while reducing the impact of disjoint and fragmented information that may occur when arbitrarily splitting articles.

\insertretrievalexperimentstoptwenty

\section{Training and Evaluation Details}

\subsection{mDPR Reader:}
We train a multilingual DPR reader model using pretrained bert-base-multilingual-uncased \footnote{\url{https://huggingface.co/bert-base-multilingual-uncased}} as the model backbone.
The model was trained to predict the correct answer span for a question given a set of relevant passages.
We trained our model using the DPR retriever output\footnote{\url{https://github.com/facebookresearch/DPR}} on the training and development set of Natural questions and evaluated on the test set of \afriqa in a zero-shot manner.
The model was trained on 4 A6000 Nvidia GPUs with a batch size of 16 and 2 gradient accumulation steps. We used an initial learning rate of 5e-5 and 500 warmup steps.
The full list of training hyperparameters can be found in \autoref{tab:modelconfig}.

\subsection{AfroXLM-R Reader}
To extract answer spans from the gold passages, we train extractive reader models on the training set of Squad 2.0~\cite{2016arXiv160605250R} and fQuad \cite{dhoffschmidt-etal-2020-fquad} using AfroXLM-R as a backbone.
We evaluated the models on the test queries and the annotated gold passages.
The models were trained for 5 epochs using a fixed learning rate of 3e-5 and batch size of 16 on a single A100 Nvidia GPU.

\subsection{mT5 Reader}

We fine-tuned multilingual pre-trained text-to-text transformer (mT5) ~\cite{xue2020mt5} on Squad 2.0~\cite{2016arXiv160605250R} dataset to generate answers from the gold passages.
We trained the model for 5 epochs with a learning rate of 3e-5 and batch size of 32 on a single A100 Nvidia GPU.

\section{Additional Experiments}

\subsection{Retrieval Top-20 Accuracy}

We present top-20 retriever accuracy results in \autoref{tab:retrievalexperimentstoptwenty}.

This further highlights the downstream effect of translation quality on retriever effectiveness with human translations showing better accuracy than other machine translation systems.

\subsection{XOR-Full Results}
\autoref{tab:fulltripembleu} presents the Exact Match Accuracy and BLEU scores of the XOR-Full task.
The table contains downstream results of different translation-retriever pipelines to extract the  answer span and translate it back to the same language as the question.

\subsection{Question Length Distribution}
Across all languages, questions are usually within 10 words with few outliers greater than 20 as shown in \autoref{fig:question_length}. While Swahili and Bemba have the tightest bound, Zulu, Yoruba, and Wolof have quite a handful of questions that extend past 20 words.

\section{Summary of Language Linguistic Properties}
In \autoref{tab:languagelinguistic}, we provide a structured breakdown of the typologies, grammatical structures, and phonology of the 10 languages in \afriqa.

\insertlanginfo

\end{document}